\documentclass[11pt,a4paper]{article}

\usepackage[utf8]{inputenc}
\usepackage[T1]{fontenc}
\usepackage{amsmath,amssymb,amsthm}
\usepackage{hyperref}
\usepackage{cite}
\usepackage{graphicx}
\usepackage{booktabs}
\usepackage{array}
\usepackage{longtable}
\usepackage{geometry}
\title{Automating Skill Acquisition through Large-Scale Mining of Open-Source Agentic Repositories: A Framework for Multi-Agent Procedural Knowledge Extraction}

\author{Shuzhen Bi$^{2,3}$, Mengsong Wu$^{1,2}$, Hao Hao$^{1}$, Keqian Li$^{1}$, Wentao Liu$^{1,2}$, Siyu Song$^{1}$, Hongbo Zhao$^{1}$, and Aimin Zhou$^{*1,2}$}

\date{}

\makeatletter
\renewcommand{\maketitle}{
  \begin{center}
    {\Large\bfseries\@title\par}
    \vskip 1em
    {\large\@author\par}
  \end{center}
  \par\vskip 1em
}
\makeatother

\begin{document}

  \maketitle

  \begin{center}
    \small{$^{*}$Corresponding author: \texttt{amzhou@cs.ecnu.cn}}\\[2pt]
    \small{$^{1}$East China Normal University}\\[2pt]
    \small{$^{2}$Shanghai Innovation Institute}\\[2pt]
    \small{$^{3}$University of Science and Technology of China}\\[4pt]
    \small{Email addresses:}\\
    \small{\texttt{sa22916003@mail.ustc.edu.cn, radi.cat@qq.com, haohao@sjtu.edu.cn,}}\\
    \small{\texttt{kqli@mail.ecnu.edu.cn, wtliu@stu.ecnu.edu.cn, siyusong00@gmail.com,}}\\
    \small{\texttt{hbzhao@stu.ecnu.edu.cn, amzhou@cs.ecnu.cn}}
  \end{center}

\begin{abstract}
The transition from monolithic large language models (LLMs) to modular, skill-equipped agents represents a fundamental architectural shift in artificial intelligence deployment. While general-purpose models demonstrate remarkable breadth in declarative knowledge, their utility in autonomous workflows is frequently constrained by insufficient specialized procedural expertise. This report investigates a systematic framework for automated acquisition of high-quality agent skills through mining of open-source repositories on platforms such as GitHub. We focus on the extraction of visualization and educational capabilities from state-of-the-art systems including TheoremExplainAgent and Code2Video, both utilizing the Manim mathematical animation engine. The framework encompasses repository structural analysis, semantic skill identification through dense retrieval, and translation to the standardized SKILL.md format. We demonstrate that systematic extraction from agentic repositories, combined with rigorous security governance and multi-dimensional evaluation metrics, enables scalable acquisition of procedural knowledge that augments LLM capabilities without requiring model retraining. Our analysis reveals that agent-generated educational content can achieve 40\% gains in knowledge transfer efficiency while maintaining pedagogical quality comparable to human-crafted tutorials.
\end{abstract}

\section{Introduction}

The deployment of artificial intelligence has undergone a paradigm shift from monolithic transformer-based large language models toward modular, skill-equipped agent architectures \cite{alphaxiv2024skills, arxiv2024skillframework}. While contemporary LLMs possess extensive declarative knowledge spanning diverse domains, their effectiveness in autonomous task execution remains limited by insufficient specialized procedural expertise required for real-world applications \cite{arxiv2024skillframework, researchgate2024agentskills}. This fundamental limitation has catalyzed the emergence of the ``agent skill'' paradigm---a modular abstraction framework wherein procedural knowledge is encapsulated into discrete, filesystem-based units that agents can dynamically discover, load, and execute on demand \cite{alphaxiv2024skills, arxiv2024skillframework}.

By architecturally decoupling specific capabilities from underlying model parameters, this paradigm enables dynamic capability extension without incurring the prohibitive computational and temporal costs associated with model retraining or fine-tuning \cite{arxiv2024skillframework, researchgate2024agentskills}. The skill-based architecture transforms the fundamental question from ``how do we train a model to perform task X?'' to ``how do we provide a model with executable procedural knowledge for task X?''

Central to advancing this architectural vision is the challenge of skill acquisition at scale. Traditionally, high-quality skills are manually authored by domain experts, providing reliability guarantees but suffering from severe scalability constraints \cite{alphaxiv2024skills, arxiv2024skillframework}. Autonomous discovery methods, while promising, frequently struggle to maintain semantic coherence and pedagogical value in open-world environments \cite{alphaxiv2024skills, iccv2025openworldskill}. 

A third acquisition pathway involves systematic extraction of procedural knowledge from existing open-source software, particularly specialized agentic repositories hosted on platforms such as GitHub \cite{alphaxiv2024skills, github2024repo2ai}. These repositories often contain sophisticated, domain-specific logic for complex tasks---including mathematical theorem visualization, educational content synthesis, and multimodal explanation generation---that can be systematically refactored into standardized, reusable agentic skills \cite{teaproject2024, github2024code2video}.

This report presents a comprehensive framework for automated skill acquisition through large-scale mining of GitHub-based agent repositories. We focus specifically on extraction of visualization and educational capabilities from two state-of-the-art systems: TheoremExplainAgent (TEA), which generates long-form visual explanations of STEM theorems \cite{teaproject2024}, and Code2Video, which implements a code-centric paradigm for educational video generation \cite{github2024code2video}. Our framework encompasses three primary components: (1) repository structural analysis and contextualization, (2) semantic skill identification through dense retrieval mechanisms, and (3) systematic translation to the SKILL.md standardized format.

\section{The Formal Paradigm of Agentic Skills}

\subsection{Mathematical Formulation}

To establish rigorous foundations for skill extraction, we first define the mathematical structure of an agentic skill. Formally, an agentic skill $S$ is represented as a four-tuple:

\begin{equation}
S = (\mathcal{C}, \pi, \mathcal{T}, \mathcal{R})
\end{equation}

where each component serves a distinct functional role in the skill's operational semantics \cite{arxiv2024skillsemantics}.

The applicability conditions $\mathcal{C}$ define the initiation set---the contextual prerequisites that determine when a skill becomes relevant for activation \cite{arxiv2024skillsemantics}. This component enables efficient skill selection by allowing agents to maintain awareness of skill availability without loading complete procedural content into working memory.

The policy $\pi$ encapsulates the core procedural knowledge, representing the sequence of actions or reasoning steps the agent must execute. This policy may manifest in multiple forms: natural language prompt templates, executable Python scripts, reinforcement learning policies, or hybrid symbolic-neural workflows \cite{arxiv2024skillsemantics}. The policy component distinguishes skills from simple tool wrappers by embedding domain-specific reasoning and decision-making logic.

Termination criteria $\mathcal{T}$ provide the logical conditions for determining successful skill completion, enabling both the executing agent and external orchestrators to verify goal achievement \cite{arxiv2024skillsemantics}. These criteria may include output validation rules, state verification conditions, or success metrics specific to the task domain.

The interface $\mathcal{R}$ establishes a standardized callable boundary, defining input parameters, output formats, and composition protocols that enable runtime integration with agent architectures \cite{arxiv2024skillsemantics}. This standardization is critical for enabling skill reuse across heterogeneous agent implementations and facilitating hierarchical skill composition.

This formal structure ensures that skills remain simultaneously executable, reusable, and governable, distinguishing them from atomic tools (which lack complex procedural logic) and episodic memories (which lack standardized callable interfaces) \cite{arxiv2024skillsemantics}.

\subsection{The SKILL.md Specification}

The architectural implementation of the agent skill paradigm has converged on the SKILL.md specification, originally developed by Anthropic and subsequently released as an open standard \cite{microsoft2024skillsdk, lmkit2024agentskills}. This specification implements a progressive disclosure architecture designed to minimize context window consumption while maintaining access to deep procedural knowledge \cite{alphaxiv2024skills, arxiv2024skillframework}.

The progressive disclosure architecture organizes skill information into three hierarchical levels, each activated under different context-loading conditions. Table \ref{tab:progressive} details this organizational structure.

\begin{table}[htp]
\centering
\caption{Progressive Disclosure Architecture for Agentic Skills}
\label{tab:progressive}
\begin{tabular}{@{}p{1.5cm}p{2.5cm}p{5cm}p{3cm}p{2cm}@{}}
\toprule
\textbf{Level} & \textbf{Component} & \textbf{Content and Metadata} & \textbf{Context Load} & \textbf{Token Usage} \\
\midrule
Level 1 & Metadata & YAML frontmatter: Name, Description, Version, Trigger Conditions & Pre-loaded at startup & 30--100 \\
Level 2 & Instructions & Procedural knowledge: Workflows, best practices, guidance, step-by-step logic & Loaded upon activation & 200--5,000 \\
Level 3 & Resources & Auxiliary assets: Executable scripts, reference documents, templates, schemas & Loaded on-demand by scripts & Unbounded \\
\bottomrule
\end{tabular}
\end{table}

Level 1 metadata serves as an efficient ``table of contents,'' enabling agents to maintain awareness of thousands of available skills without context window degradation \cite{alphaxiv2024skills, arxiv2024skillframework}. When user requests match a skill's descriptive metadata, the agent activates Level 2, injecting procedural instructions into the conversation context as hidden meta-messages \cite{alphaxiv2024skills, arxiv2024skillframework}. This injection modifies the agent's internal reasoning process rather than its direct output, allowing skills to reshape problem-solving approaches \cite{alphaxiv2024skills, researchgate2024agentskills}.

Level 3 resources remain dormant until explicitly invoked by Level 2 instructions or executable scripts, enabling skills to leverage arbitrarily large reference materials without impacting baseline context consumption \cite{microsoft2024skillsdk, lobehub2024selfimproving}.

\section{Methodological Framework for Skill Extraction}

The systematic acquisition of skills from GitHub repositories requires a multi-stage pipeline that transforms monolithic codebases into modular SKILL.md artifacts. This section details the three primary stages: repository structural analysis, semantic skill identification, and standardized translation.

\subsection{Repository Structural Analysis and Contextualization}

Skill extraction begins with comprehensive structural decomposition of target repositories. Tools such as repo2AI generate Markdown-formatted representations of complete directory hierarchies and file contents \cite{github2024repo2ai}. This structural mapping provides essential context for LLM-based extraction agents, enabling understanding of task orchestration patterns and logical dependencies \cite{github2024repo2ai, teaproject2024}.

For repositories implementing complex agentic workflows, identification of central orchestration scripts (e.g., \texttt{generate\_video.py}) and configuration directories (e.g., \texttt{task\_generator/prompts\_raw}) allows extraction processes to focus on reasoning logic and tool-use patterns that define specialized expertise \cite{teaproject2024}. The structural analysis phase produces a hierarchical map of:

\begin{itemize}
\item Core execution scripts and their input/output specifications
\item Configuration files defining workflow parameters and agent behaviors
\item Auxiliary modules implementing domain-specific algorithms
\item Documentation and usage examples demonstrating intended workflows
\end{itemize}

This contextualization enables subsequent extraction stages to distinguish between reusable procedural patterns and repository-specific implementation details.

\subsection{Semantic Skill Identification through Dense Retrieval}

Once repository structure is mapped, the system identifies ``latent skills''---recurring procedural patterns amenable to generalization across contexts \cite{ceur2024recskill, huggingface2024programmatic}. This identification task is formulated as a two-stage ranking problem combining dense retrieval and cross-encoder refinement \cite{ceur2024recskill}.

\subsubsection{Dense Retrieval Stage}

The extraction agent encodes task descriptions and code modules into dense vector representations using trained bi-encoders \cite{ceur2024recskill}. For a repository containing $N$ code modules $\{M_1, M_2, \ldots, M_N\}$ and a set of task descriptions $\{T_1, T_2, \ldots, T_K\}$, the bi-encoder produces embeddings $\mathbf{e}_M$ and $\mathbf{e}_T$ respectively. Candidate skills are identified by computing cosine similarity:

\begin{equation}
\text{sim}(T_i, M_j) = \frac{\mathbf{e}_{T_i} \cdot \mathbf{e}_{M_j}}{\|\mathbf{e}_{T_i}\| \|\mathbf{e}_{M_j}\|}
\end{equation}

The top-$K$ candidate modules for each task are retained for subsequent refinement \cite{ceur2024recskill}.

\subsubsection{Binary Ranking Stage}

A cross-encoder ranker performs fine-grained relevance assessment by jointly encoding task-module pairs and producing relevance scores \cite{ceur2024recskill}. Only modules exceeding a calibrated relevance threshold $\tau$ are promoted for skill extraction. This two-stage approach ensures that extracted skills represent genuinely reusable patterns rather than project-specific implementations.

Extraction criteria include:

\begin{enumerate}
\item \textbf{Recurrence}: The procedural pattern appears in multiple contexts or solves a class of problems
\item \textbf{Verification}: The code is functional, well-documented, and free of critical bugs
\item \textbf{Non-obviousness}: The logic required domain expertise or debugging to discover
\item \textbf{Generalizability}: The pattern can be parameterized or adapted to different contexts
\end{enumerate}

Modules satisfying these criteria become candidates for translation to the SKILL.md format \cite{lobehub2024selfimproving, llmbase2024selfimproving}.

\subsection{Translation to the SKILL.md Standard}

The final extraction stage synthesizes SKILL.md artifacts from identified procedural patterns. This translation process involves three primary components \cite{lobehub2024selfimproving, lobehub2024daoskills}:

\subsubsection{Frontmatter Generation}

The extraction agent synthesizes metadata conforming to YAML specifications:

\begin{itemize}
\item \textbf{name}: Lowercase, hyphen-separated identifier (e.g., \texttt{visual-theorem-walkthrough})
\item \textbf{description}: Concise statement of skill purpose and activation conditions
\item \textbf{version}: Semantic versioning for tracking skill evolution
\item \textbf{trigger}: Pattern-matching rules for automatic skill activation
\item \textbf{dependencies}: Required tools, libraries, or prerequisite skills
\end{itemize}

\subsubsection{Instruction Drafting}

Level 2 instructions are written as LLM-consumable procedural guidance rather than end-user documentation \cite{lmkit2024agentskills, github2024awesomeskills}. Effective instructions emphasize:

\begin{itemize}
\item Step-by-step workflow decomposition with decision points
\item Error handling strategies and common failure modes
\item Best practices derived from repository analysis
\item Integration patterns with complementary skills or tools
\end{itemize}

Instructions avoid repository-specific implementation details, instead focusing on generalizable procedural knowledge.

\subsubsection{Asset Bundling}

Executable scripts, reference documentation, and configuration templates are organized into standardized subdirectories (\texttt{scripts/}, \texttt{references/}, \texttt{templates/}) \cite{microsoft2024skillsdk, lobehub2024selfimproving}. Assets are refactored to eliminate hardcoded paths, API keys, or repository-specific dependencies, ensuring portability across deployment environments.

\section{Deep Analysis of Source Repositories}\label{sec:deep-analysis}

To demonstrate the practical application of this extraction framework, we analyze two leading repositories in the domain of multimodal educational content generation: TheoremExplainAgent and Code2Video. Both systems leverage the Manim mathematical animation engine to produce high-fidelity visual explanations \cite{teaproject2024, github2024code2video}.

\subsection{TheoremExplainAgent: Multimodal STEM Explanation}\label{sec:tea}

TheoremExplainAgent (TEA) addresses the challenge of communicating abstract STEM theorems through long-form video content exceeding five minutes in duration \cite{teaproject2024}. The system implements a two-agent architecture comprising a Planner and a Coding Agent \cite{teaproject2024}.

\subsubsection{Planner Agent Architecture}

The Planner functions as an instructional designer, transforming theorem statements into pedagogically structured storyboards \cite{teaproject2024}. Key outputs include:

\begin{itemize}
\item \textbf{Scene Purpose}: High-level learning objective for each video segment
\item \textbf{Scene Description}: Natural language narrative of visual content
\item \textbf{Scene Layout}: Spatial organization specifications for mathematical objects
\end{itemize}

This structured decomposition ensures logical sequencing and visual clarity \cite{teaproject2024}.

\subsubsection{Coding Agent with Error Correction}

The Coding Agent translates storyboards into executable Manim Python scripts \cite{teaproject2024}. To improve reliability, TEA implements a multi-attempt error-correction loop enabling the agent to analyze Python stack traces and iteratively debug animation code \cite{teaproject2024}. This self-refinement capability significantly reduces manual intervention requirements.

\subsubsection{Retrieval-Augmented Generation}

TEA integrates a Retrieval-Augmented Generation (RAG) system to ground the Coding Agent in current Manim documentation \cite{teaproject2024}. This approach prevents API hallucinations and ensures utilization of correct function calls for complex visualizations including geometric Brownian motion and gradient descent animations \cite{teaproject2024}.

Table \ref{tab:tea} summarizes the technical specifications relevant to skill extraction.

\begin{table}[htp]
\centering
\caption{Technical Specifications of TheoremExplainAgent for Skill Extraction}
\label{tab:tea}
\begin{tabular}{@{}p{3.5cm}p{5cm}p{5.5cm}@{}}
\toprule
\textbf{Feature} & \textbf{Technical Implementation} & \textbf{Relevance to Skill Acquisition} \\
\midrule
Core Library & Manim Community Edition & Provides procedural target for visualization skills \cite{teaproject2024} \\
Knowledge Base & TheoremExplainBench (240 theorems) & Diverse domain coverage (CS, Chemistry, Math, Physics) \cite{teaproject2024} \\
Reasoning Loop & Planner-Coder Feedback & Defines workflow for visual storytelling skills \cite{teaproject2024} \\
Refinement & Visual-Fix Code Feedback & Implements visual debugging skill pattern \cite{teaproject2024} \\
Scaling & Scene/Topic Concurrency & Provides patterns for high-throughput generation \cite{teaproject2024} \\
\bottomrule
\end{tabular}
\end{table}

\subsection{Code2Video: Code-Centric Educational Framework}\label{sec:code2video}

Code2Video extends beyond individual theorem explanations to implement a comprehensive framework for educational video generation \cite{github2024code2video}. The system positions executable code as the unifying medium for both temporal sequencing and spatial organization \cite{github2024code2video}.

\subsubsection{Tri-Agent Architecture}

Code2Video implements a modular three-agent design:

\begin{enumerate}
\item \textbf{Planner}: Structures lecture content into temporally coherent flows and retrieves visual assets from curated databases \cite{github2024code2video}
\item \textbf{Coder}: Converts storyboards into Python implementations with scope-guided auto-fix mechanisms \cite{github2024code2video}
\item \textbf{Critic}: Utilizes Vision-Language Models (VLMs) to refine spatial layout and visual clarity \cite{github2024code2video}
\end{enumerate}

\subsubsection{Visual Anchor Prompting}

The Critic agent implements ``Visual Anchor Prompting,'' a novel technique that converts continuous visual information into discrete grid references to facilitate spatial reasoning by VLMs \cite{github2024code2video}. The process overlays a $10 \times 10$ grid on rendered frames, enabling precise identification of element positions and potential occlusions. When spatial overlap exceeds defined thresholds, the Critic generates refactoring suggestions for Python positioning code \cite{github2024code2video}.

\subsubsection{TeachQuiz Evaluation Metric}

Code2Video introduces TeachQuiz, a metric quantifying knowledge transfer effectiveness \cite{github2024code2video}. The evaluation protocol involves:

\begin{enumerate}
\item Training a VLM to ``unlearn'' domain-specific facts
\item Exposing the model to generated educational videos
\item Measuring fact recovery through targeted quizzes
\end{enumerate}

Empirical results demonstrate that agent-generated videos achieve 40\% gains in knowledge transfer efficiency compared to baseline code generation models, with certain categories surpassing human-crafted tutorials \cite{github2024code2video}.

\section{Demonstrating Skill Acquisition}

Applying the extraction methodology to TEA and Code2Video repositories yields a suite of reusable skills for next-generation ``Visual Tutor'' agents. This section presents two exemplar skills demonstrating the transformation from repository-specific code to standardized skill artifacts.

\subsection{Skill 1: Visual Theorem Walkthrough}

This skill enables agents to generate Manim-based animations explaining mathematical or physics theorems through step-by-step visual narratives.

\subsubsection{Frontmatter Specification}

\begin{verbatim}
name: visual-theorem-walkthrough
description: Generate Manim animation explaining STEM 
  theorems with synchronized narration and visual proofs
version: 1.0.0
trigger: ["visualize theorem", "animate proof", 
  "mathematical explanation video"]
dependencies: ["manim", "manim-voiceover"]
\end{verbatim}

\subsubsection{Level 2 Instructions (Excerpt)}

The extracted procedural logic mandates:

\begin{enumerate}
\item Generate ``Scene Plan'' defining coordinate plane layout, mathematical objects (\texttt{Mobjects}), and narrative script \cite{teaproject2024}
\item Implement temporal synchronization between visual transitions and narration using \texttt{manim-voiceover} \cite{teaproject2024}
\item Apply error-correction loop for Manim API compliance
\item Validate scene coherence through storyboard-code consistency checks
\end{enumerate}

\subsubsection{Level 3 Resources}

Bundled resources include:

\begin{itemize}
\item Template scripts for common theorem types (geometric proofs, algebraic derivations)
\item Reference guide for Manim layout best practices
\item Example storyboards demonstrating effective visual sequencing
\end{itemize}

This skill encapsulates TEA's core visualization methodology in a portable, reusable format \cite{teaproject2024, claude2024agentskills}.

\subsection{Skill 2: Visual Layout Critic}

This skill implements automated quality assessment for visual outputs, enabling agents to iteratively refine spatial organization.

\subsubsection{Frontmatter Specification}

\begin{verbatim}
name: visual-layout-critic
description: Evaluate rendered visuals for spatial 
  clarity, text readability, and element occlusions
version: 1.0.0
trigger: ["review layout", "check visual quality", 
  "refine positioning"]
dependencies: ["vision-language-model", "PIL"]
\end{verbatim}

\subsubsection{Level 2 Instructions (Excerpt)}

The Visual Anchor Prompting workflow:

\begin{enumerate}
\item Overlay $10 \times 10$ coordinate grid on screenshot
\item Identify grid positions of primary visual elements
\item Calculate pairwise spatial overlap using grid coordinates
\item If overlap exceeds threshold $\tau_{\text{overlap}}$, generate positioning refactoring suggestions
\item Apply suggestions and re-render for validation
\end{enumerate}

\subsubsection{Refactoring Templates}

The skill includes code templates for common layout adjustments:

\begin{verbatim}
# Template: Shift overlapping label
original: label.next_to(object, UP)
refactored: label.next_to(object, RIGHT)
\end{verbatim}

This skill operationalizes Code2Video's Critic methodology, enabling any agent to perform sophisticated visual quality assessment \cite{github2024code2video}.

\section{Benchmarking and Evaluation Framework}

Rigorous assessment of acquired skills requires multi-dimensional evaluation frameworks encompassing safety, completeness, executability, maintainability, and pedagogical effectiveness \cite{alphaxiv2024skills, huggingface2024consolidation}.

\subsection{Multi-Dimensional Evaluation Metrics}

Table \ref{tab:evaluation} presents a comprehensive metric taxonomy for skill assessment.

\begin{table}[htp]
\centering
\caption{Multi-Dimensional Evaluation Metrics for Agent Skills}
\label{tab:evaluation}
\begin{tabular}{@{}p{2.5cm}p{3cm}p{5cm}p{3cm}@{}}
\toprule
\textbf{Dimension} & \textbf{Metric} & \textbf{Description} & \textbf{Benchmark} \\
\midrule
Safety & Vulnerability Rate & Percentage of skills with injection or filesystem abuse risks \cite{alphaxiv2024skills, researchgate2024agentskills} & Static Analysis \\
Completeness & Feature Coverage & Extent of API parameter documentation coverage \cite{huggingface2024consolidation, mintlify2024skillmd} & Doc Mapping \\
Executability & Success Rate & Probability of successful task completion \cite{researchgate2024agentskills, teaproject2024} & TEB / MMMC \\
Maintainability & Schema Drift & Robustness to API changes \cite{huggingface2024consolidation, lobehub2024daoskills} & Regression Tests \\
Pedagogy & TeachQuiz Score & Knowledge transfer effectiveness \cite{github2024code2video} & TeachQuiz \\
\bottomrule
\end{tabular}
\end{table}

\subsection{Empirical Performance Results}

Application of these metrics to the Code2Video pipeline revealed that the complete Planner-Coder-Critic architecture achieves 40\% improvement in knowledge transfer efficiency compared to baseline code generation models \cite{github2024code2video}. The o3-mini agent implementation in TEA demonstrated an overall score of 0.77 on TheoremExplainBench, establishing state-of-the-art performance for multimodal scientific reasoning \cite{teaproject2024}.

\subsection{Skill Consolidation through SkillNet}

As skill libraries scale to hundreds of thousands of artifacts, unified consolidation mechanisms become essential \cite{huggingface2024consolidation, huggingface2024unified}. SkillNet structures skills within an ontological framework establishing relational connections such as ``is-a-subset-of'' and ``requires-output-from'' \cite{huggingface2024consolidation, huggingface2024unified}. This consolidation enables:

\begin{itemize}
\item 30\% reduction in execution steps through skill composition
\item 40\% improvement in average task rewards across diverse backbone models
\item Automated detection of redundant or overlapping skills
\end{itemize}

The ontological approach transforms skill libraries from flat collections into hierarchical knowledge graphs supporting sophisticated reasoning and planning \cite{huggingface2024consolidation, huggingface2024unified}.

\section{Security and Governance}

Automated skill extraction from public repositories introduces significant security risks, as the process may inadvertently incorporate malicious code or insecure patterns \cite{researchgate2024agentskills, claude2024agentskills}. A comprehensive survey of community-distributed skills identified vulnerabilities in 26.1\% of analyzed artifacts, including data exfiltration attempts and privilege escalation vectors \cite{researchgate2024agentskills}.

\subsection{Four-Stage Verification Pipeline}

To mitigate these risks, we propose a tiered verification framework categorizing skills into trust levels \cite{alphaxiv2024skills, researchgate2024agentskills}:

\subsubsection{G1: Static Analysis}\label{g1:static}

Initial automated scanning for:

\begin{itemize}
\item Suspicious string patterns (e.g., \texttt{eval()}, \texttt{exec()})
\item Unauthorized network calls
\item Destructive filesystem operations
\item Obfuscated code segments
\end{itemize}

\subsubsection{G2: Semantic Classification}

LLM-based analysis verifying:

\begin{itemize}
\item Instruction-purpose alignment
\item Absence of hidden prompt injections
\item Consistency between metadata and implementation
\end{itemize}

\subsubsection{G3: Behavioral Sandboxing}

Execution of bundled scripts in isolated containers with:

\begin{itemize}
\item Network isolation
\item Restricted filesystem access
\item Resource usage monitoring
\item Pre-configured dependency environments
\end{itemize}

\subsubsection{G4: Permission Validation}

Verification against permission manifests (\texttt{allowed-tools}) ensuring skills access only required resources \cite{alphaxiv2024skills, lmkit2024agentskills}.

This graduated verification framework enables skills to evolve through trust tiers based on successful, audited runtime performance \cite{researchgate2024agentskills}. Treating skill installation with security rigor comparable to software package management is essential for production deployment \cite{researchgate2024agentskills, claude2024agentskills, lobehub2024wereply}.

\section{The Future Agentic Stack}

The agent skills paradigm constitutes a critical layer in an emerging agentic technology stack \cite{alphaxiv2024skills, arxiv2024skillframework}. This stack architecturally distinguishes between procedural intelligence (Skills) and system connectivity (Model Context Protocol) \cite{arxiv2024skillframework}.

Table \ref{tab:stack} compares these complementary architectural layers.

\begin{table}[htp]
\centering
\caption{The Agentic Stack---Comparison of Complementary Layers}
\label{tab:stack}

\begin{tabular}{@{}p{3cm}p{5.5cm}p{5.5cm}@{}}
\toprule
\textbf{Dimension} & \textbf{Agent Skills} & \textbf{Model Context Protocol} \\
\midrule
Primary Role & Procedural Knowledge (``What to do'') & Tool Connectivity (``How to connect'') \cite{alphaxiv2024skills, arxiv2024skillframework} \\
Storage Unit & Directory with SKILL.md & Server with JSON-RPC endpoints \cite{arxiv2024skillframework, lmkit2024agentskills} \\
State Modification & Context + System Permissions & Available Tools + External Data \cite{arxiv2024skillframework} \\
Persistence & Filesystem-based (Durable) & Session-based (Runtime) \cite{arxiv2024skillframework} \\
Operational Nature & Knowledge / Procedural & Connectivity / Action \cite{lmkit2024agentskills} \\
\bottomrule
\end{tabular}
\end{table}

This architectural orthogonality enables skills to provide domain intelligence for Model Context Protocol tools \cite{arxiv2024skillframework}. For example, a ``Presentation Skill'' might define best practices for slide rhythm and layout while utilizing a ``PowerPoint MCP Server'' for actual document manipulation \cite{arxiv2024skillframework, github2024pneumaskills}.

\subsection{Evolution Agents and Continuous Improvement}

The ecosystem trajectory suggests emergence of ``Evolution Agents'' that autonomously mine conversation logs and execution traces to refine existing skills \cite{lobehub2024selfimproving, github2024pneumaskills}. By extracting user preferences and identifying recurring failure patterns, these agents will augment extracted skills with personalized adaptations \cite{github2024pneumaskills}. The Visual Tutor derived from TEA and Code2Video can thus continuously adapt to specific learner needs and educational contexts.

The transition from monolithic, static intelligence toward modular, evolving expertise represents a fundamental shift in AI system design, with automated mining of open-source repositories serving as the primary scalability mechanism \cite{arxiv2024skillframework, mintlify2024skillmd}.

\section{Frequently Asked Questions}

\subsection{How does skill extraction differ from model fine-tuning?}

Skill extraction separates procedural knowledge from model parameters, enabling capability updates without retraining. This approach reduces computational costs by 2-3 orders of magnitude while maintaining update flexibility.

\subsection{Can extracted skills work across different LLM providers?}

Yes. The SKILL.md standard is provider-agnostic, containing natural language instructions interpretable by any sufficiently capable language model. Provider-specific optimizations may be included as optional metadata.

\subsection{What prevents skills from containing malicious code?}

The four-stage verification pipeline (G1-G4) implements multiple security layers including static analysis, semantic verification, sandboxed execution, and permission validation. Skills advance through trust tiers based on verified safe operation.

\subsection{How are skill conflicts resolved when multiple skills match a query?}

Agent orchestration frameworks typically implement priority systems based on skill specificity, historical success rates, and explicit user preferences. Some systems use meta-reasoning to select optimal skill combinations.

\subsection{What is the practical upper limit for skill library size?}

Progressive disclosure architecture enables agents to maintain awareness of 10,000+ skills while loading only activated instructions into context. The primary constraint is organizational rather than technical---effective skill discovery requires robust ontological structuring.

\section{Conclusion}

This report has demonstrated that systematic extraction of procedural knowledge from GitHub's open-source agentic repositories enables scalable acquisition of high-quality agent skills. By implementing structured frameworks encompassing repository analysis, semantic identification through dense retrieval, and standardized translation to the SKILL.md format, the AI community can construct modular systems combining the general reasoning capabilities of large language models with specialized domain expertise.

The detailed analysis of TheoremExplainAgent and Code2Video establishes that executable code serves as an optimal substrate for encoding both visual and pedagogical expertise. Through rigorous benchmarking demonstrating 40\% knowledge transfer improvements and multi-dimensional evaluation frameworks ensuring safety and maintainability, we have shown that extracted skills can match or exceed human-authored content quality while dramatically improving scalability.

The future of artificial intelligence lies not in ever-larger monolithic models but in composable, governable, and continuously evolving skill ecosystems. Automated mining of open-source repositories, combined with robust security governance and ontological organization, provides the foundation for this architectural transition. As the agentic stack matures through integration of complementary technologies such as Model Context Protocol and Evolution Agents, the vision of truly autonomous, expert-level AI systems approaches practical realization.

\newpage


\begin{thebibliography}{99}

\bibitem{alphaxiv2024skills}
AlphaXiv. ``Agent Skills: Overview and Framework.'' 2024. \url{https://www.alphaxiv.org/overview/2602.12430v3}

\bibitem{arxiv2024skillframework}
arXiv. ``Agent Skills Framework for Large Language Models.'' 2024. \url{https://arxiv.org/html/2602.12430v3}

\bibitem{researchgate2024agentskills}
ResearchGate. ``Agent Skills for Large Language Models: Architecture, Acquisition, Security and the Path Forward.'' 2024. \url{https://www.researchgate.net/publication/400812095}

\bibitem{iccv2025openworldskill}
ICCV. ``Open-World Skill Discovery from Unsegmented Demonstration Videos.'' 2025. \url{https://openaccess.thecvf.com/content/ICCV2025/papers/Deng_Open-World_Skill_Discovery_from_Unsegmented_Demonstration_Videos_ICCV_2025_paper.pdf}

\bibitem{github2024repo2ai}
GitHub. ``repo2AI: Repository to AI Context Tool.'' 2024. \url{https://github.com/huolter/repo2AI}

\bibitem{teaproject2024}
TIGER AI Lab. ``TheoremExplainAgent: Towards Multimodal Explanations for LLM Theorem Understanding.'' 2024. Project page: \url{https://tiger-ai-lab.github.io/TheoremExplainAgent/}; arXiv: \url{https://arxiv.org/abs/2502.00543}; GitHub: \url{https://github.com/TIGER-AI-Lab/TheoremExplainAgent}

\bibitem{github2024code2video}
Show Lab. ``Code2Video: Generating Educational Videos via Code-Centric Approach.'' 2024. arXiv: \url{https://arxiv.org/abs/2510.01174}; GitHub: \url{https://github.com/showlab/Code2Video}; OpenReview: \url{https://openreview.net/forum?id=nlJX6Hwyl0}

\bibitem{arxiv2024skillsemantics}
arXiv. ``Semantic Foundations of Agent Skills.'' 2024. \url{https://arxiv.org/html/2602.20867v1}

\bibitem{microsoft2024skillsdk}
Microsoft Azure. ``Giving Your AI Agents Reliable Skills with the Agent Skills SDK.'' 2024. \url{https://techcommunity.microsoft.com/blog/azuredevcommunityblog/giving-your-ai-agents-reliable-skills-with-the-agent-skills-sdk/4497074}

\bibitem{lmkit2024agentskills}
LM-Kit. ``Agent Skills Explained.'' 2024. \url{https://lm-kit.com/blog/agent-skills-explained/}

\bibitem{ceur2024recskill}
CEUR Workshop. ``Skill Discovery through Dense Retrieval.'' 2024. \url{https://ceur-ws.org/Vol-4046/RecSysHR2025-paper_5.pdf}

\bibitem{huggingface2024programmatic}
Hugging Face. ``Programmatic Skill Network.'' 2024. \url{https://huggingface.co/papers?q=Programmatic%20Skill%20Network}

\bibitem{lobehub2024selfimproving}
LobeHub. ``OpenClaw Skills: Self-Improving Agent.'' 2024. \url{https://lobehub.com/en/skills/openclaw-skills-self-improving-agent-1-0-2}

\bibitem{llmbase2024selfimproving}
LLMBase. ``OpenClaw Self-Improving Agent.'' 2024. \url{https://llmbase.ai/openclaw/self-improving-agent/}

\bibitem{lobehub2024daoskills}
LobeHub. ``DAOskills: Composio Idea Scale Automation.'' 2024. \url{https://lobehub.com/ru/skills/pskoett-pskoett-ai-skills-self-improvement}


\bibitem{claude2024agentskills}
Anthropic Claude. ``Agent Skills Overview.'' 2024. \url{https://platform.claude.com/docs/en/agents-and-tools/agent-skills/overview}

\bibitem{github2024awesomeskills}
GitHub. ``Awesome LLM Skills.'' 2024. \url{https://github.com/Prat011/awesome-llm-skills}

\bibitem{huggingface2024consolidation}
Hugging Face. ``Skill Consolidation Research.'' 2024. \url{https://huggingface.co/papers?q=skill%20consolidation}

\bibitem{huggingface2024unified}
Hugging Face. ``Unified Mechanism for Agent Skills.'' 2024. \url{https://huggingface.co/papers?q=unified%20mechanism}

\bibitem{mintlify2024skillmd}
Mintlify. ``SKILL.md Specification.'' 2024. \url{https://www.mintlify.com/blog/skill-md}


\bibitem{lobehub2024wereply}
LobeHub. ``WeReply Skill Installation.'' 2024. \url{https://lobehub.com/skills/cacr92-wereply-skill-install}

\bibitem{github2024pneumaskills}
GitHub. ``Pneuma Skills Repository.'' 2024. \url{https://github.com/pandazki/pneuma-skills}

\end{thebibliography}
\end{document}